\documentclass{article}

\usepackage{microtype}
\usepackage{graphicx}
\usepackage{subfigure}
\usepackage{booktabs} 
\usepackage{xcolor}

\usepackage{multirow}
\usepackage{tikz}

\usepackage{hyperref}
\usepackage{bm}
\usepackage{anyfontsize}
\usepackage{multirow}
\usepackage{color, colortbl}
\definecolor{Gray}{gray}{0.9}

\newcommand{\tikzxmark}{%
\tikz[scale=0.23] {
    \draw[line width=0.7,line cap=round] (0,0) to [bend left=6] (1,1);
    \draw[line width=0.7,line cap=round] (0.2,0.95) to [bend right=3] (0.8,0.05);
}}
\newcommand{\tikzcmark}{%
\tikz[scale=0.23] {
\draw[line width=0.7,line cap=round] (0.25,0) to [bend left=10] (1,1);
    \draw[line width=0.8,line cap=round] (0,0.35) to [bend right=1] (0.23,0);
}}

\newcommand{\methodname}{InTEnt}
\newcommand{\eg}{\textit{e.g.}}
\newcommand{\ie}{\textit{i.e.}}

\usepackage[accepted]{icml2024}

\usepackage{amsmath}
\usepackage{amssymb}
\usepackage{mathtools}
\usepackage{amsthm}

\usepackage[capitalize,noabbrev]{cleveref}

\theoremstyle{plain}

\theoremstyle{definition}

\theoremstyle{remark}

\usepackage[textsize=tiny]{todonotes}

\icmltitlerunning{Segmentation with Integrated Entropy Weighting for Single Image Test-Time Adaptation}

\begin{document}

\twocolumn[
\icmltitle{Medical Image Segmentation with \methodname: Integrated Entropy \\ Weighting for Single Image Test-Time Adaptation}



\icmlsetsymbol{equal}{*}

\begin{icmlauthorlist}
\icmlauthor{Haoyu Dong}{yyy}
\icmlauthor{Nicholas Konz}{yyy}
\icmlauthor{Hanxue Gu}{yyy}
\icmlauthor{Maciej A. Mazurowski}{yyy,rad,cs,bb}
\end{icmlauthorlist}

\icmlaffiliation{yyy}{Department of Electrical and Computer Engineering, Duke University, NC, USA}
\icmlaffiliation{rad}{Department of Radiology}
\icmlaffiliation{cs}{Department of Computer Science}
\icmlaffiliation{bb}{Department of Biostatistics \& Bioinfomatics}

\icmlcorrespondingauthor{Haoyu Dong}{haoyu.dong151@duke.edu}

\icmlkeywords{Machine Learning, ICML}

\vskip 0.3in
]



\printAffiliationsAndNotice{}  

\begin{abstract}

Test-time adaptation (TTA) refers to adapting a trained model to a new domain during testing.
Existing TTA techniques rely on having multiple test images from the same domain, yet this may be impractical in real-world applications such as medical imaging, where data acquisition is expensive and imaging conditions vary frequently.
Here, we approach such a task, of adapting a medical image segmentation model with only a single unlabeled test image.
Most TTA approaches, which directly minimize the entropy of predictions, fail to improve performance significantly in this setting, in which we also observe the choice of batch normalization (BN) layer statistics to be a highly important yet unstable factor due to only having a single test domain example.
To overcome this, we propose to instead \textit{integrate} over predictions made with various estimates of target domain statistics between the training and test statistics, weighted based on their entropy statistics.
Our method, validated on $24$ source/target domain splits across $3$ medical image datasets surpasses the leading method by 2.9\% Dice coefficient on average. 
The code is available at \url{https://github.com/mazurowski-lab/single-image-test-time-adaptation}.
\end{abstract}

\section{Introduction}
Deep neural networks have demonstrated impressive performance when source (training) and target (test) images are drawn from the same distribution. 
Unfortunately, this assumption often fails in real-world applications, where target data may be corrupted naturally (\eg, with weather changes or sensor degradation \cite{Koh2020WILDSAB}) or acquired differently (\eg, MRIs taken with different scanners or under different protocols \cite{konz2023reverse}). 
Trained models can be sensitive to these shifts, resulting in performance degradation, known as the \textit{domain shift} problem \cite{QuioneroCandela2009DatasetSI, guan2021domain}.


\begin{table}[t]
\caption{Average Dice similarity coefficient (repeated 10 times, average of all source/target domain splits) of different TTA methods on three datasets. The leading performance is highlighted. \textbf{Our method outperforms the SOTA on two datasets and surpasses the leading method by $2.9\%$ on average.}
}
\vskip 0.15in
\begin{center}
\begin{small}
\begin{sc}
\begin{tabular}{l|ccc|c }
    Method &  SC  & Che. & Ret. & Avg. $\uparrow$\\
    \hline
    UNet & $57.8$ & $82.9$ & $53.3$ & $64.0$ \\
    MEMO & $59.1$ & $85.3$ & $54.1$ & $65.5$\\
    TEnt & $57.7$ & $93.0$ & $58.9$ & $68.7$\\
    SAR  & $57.5$ & $93.0$ & $58.9$ & $68.4$\\
    FSeg & $57.8$ & $93.1$ & $\mathbf{58.9}$ & $68.7$\\
    SITA & $61.3$ & $90.5$ & $56.7$ & $68.7$\\
    \hline
    InTEnt & $\mathbf{64.5}$ & $\mathbf{94.1}$ & $58.6$ & $\mathbf{71.6}$\\
\end{tabular}
\end{sc}
\end{small}
\end{center}
\vskip -0.1in
\label{tab:overall}
\end{table}

\begin{table*}[t]
\caption{\textbf{Comparison of different TTA settings and the data available in each.} 
$X^t$ and $X^s$ refer to a batch of images from the target and source domains, respectively. $x_i^t$ refers to a single image from the target domain.
``Online'' refers to whether information from prior test images is accessible for a new prediction.}
\vskip 0.15in
\begin{center}
\begin{small}
\begin{sc}
\begin{tabular}{lccccc}
\hline
Setting & Source data & Target data & Train Objective & Test Objective & Online\\
\hline
Fine-tuning & - & $X^t, Y^t$ & $L(X^t, Y^t)$ & - & \tikzcmark\\
Test-time training & $X^s, Y^s$ & $X^t$ & $L(X^t, Y^t)\!+\!L(X^s, X^t)$ & - & \tikzcmark\\ 
Test-time adaptation (TTA) & - & $X^t$ & - & $L(X^t)$ & \tikzcmark \\
Continual TTA& - & $x^t_i$ & - & $L(x_i^t)$ & \tikzcmark \\
Single Image TTA (ours) & - & $x_i^t$ & - & $L(x_i^t)$ & \tikzxmark \\
\hline
\end{tabular}
\end{sc}
\end{small}
\end{center}
\vskip -0.1in
\label{tab:setting}
\end{table*}

Early work \cite{Sun2019TestTimeTW} solves this problem by learning auxiliary tasks during training, which can be sub-optimal since the training pipeline is altered.
Fully Test-time Adaptation (TTA) methods instead propose to adapt models solely using target domain data and have achieved significant improvements in robustness to domain shift \cite{Wang2021TentFT, Ma2022TesttimeAW, Liu2021AdaptingOS}. Typically, model parameters are updated to minimize the entropy of model predictions on test images, as a proxy for minimizing the cross-entropy given that the target labels are unknown \cite{Wang2021TentFT}.
However, recent works have observed that these improvements have occurred only within certain conditions, namely that target images (1) are available in a relatively large quantity and (2) can arrive continuously, \ie, in an online fashion \cite{niu2023sharpness, gan2023decorate, zhao2023on}. The first condition further implicitly assumes that all images in the same batch are from the same domain and have balanced class information, and the second condition unavoidably favors target data that arrived later.
Both conditions bring restrictions to real-world usage.

In this paper, we consider an extreme case of TTA, where a model only has access to \textbf{a single target image} during adaptation. 
This setting is called \textbf{S}ingle \textbf{I}mage \textbf{T}est-\textbf{T}ime \textbf{A}daptation (SITTA, or SITA \cite{Khurana2021SITASI}), which we summarize and compare to related settings in Table \ref{tab:setting}. SITTA avoids the above-mentioned assumption naturally and is especially relevant to medical image analysis, where obtaining additional images from the same domain can be expensive, time-consuming, or even infeasible due to medical image privacy concerns and scanner setting inhomogeneities \cite{guan2021domain, liu2018applications}.
We focus on segmentation because it is a common yet challenging task in medical image analysis, but our method is designed generally and could be applied to other models, tasks, or applications.

\begin{figure*}[t]
    \vskip 0.2in
    \centering
    \includegraphics[width=\linewidth]{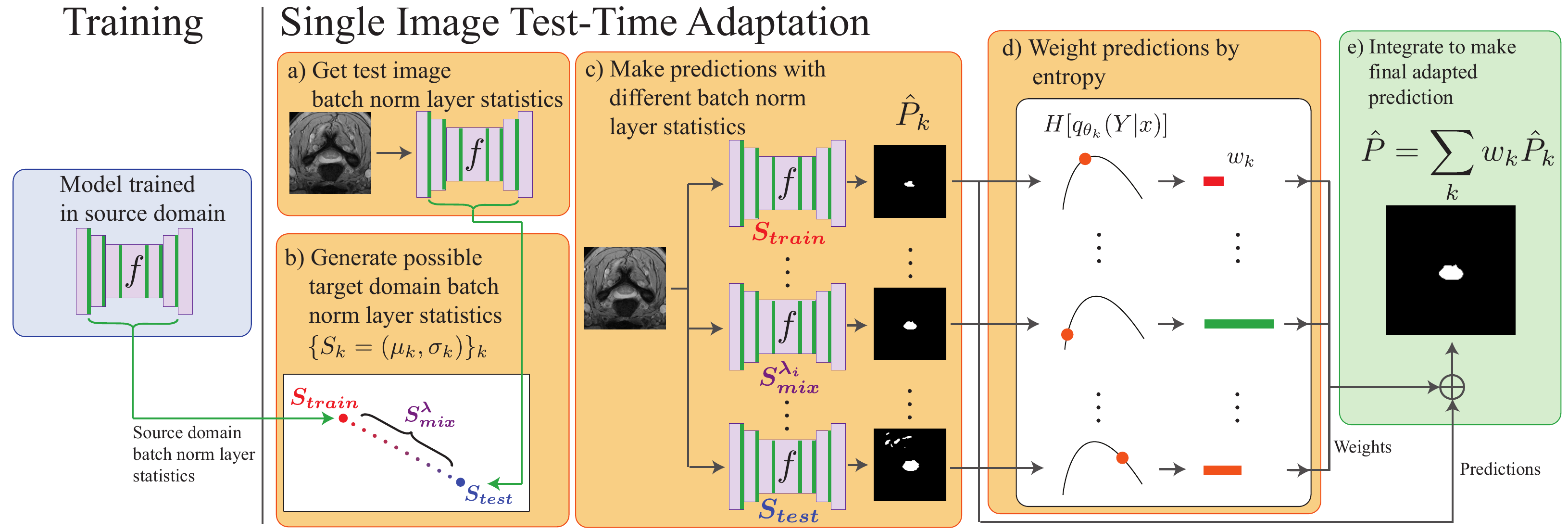}
    \caption{\textbf{Summary of our method for single-image test time adaptation of a segmentation model (Algorithm \ref{alg:main}).} Note that segmentation probability map predictions $\hat{P}_k$ and $\hat{P}$ are rounded to binary masks for visualization.}
    \label{fig:summary}
    \vskip -0.2in
\end{figure*}

After conducting many experiments within this setting, we observed that existing TTA methods, which typically optimize learnable batch normalization layer parameters (scale and shift) for the target domain, fail to alter network performance significantly. 
Instead, we find that batch norm. layer \textit{statistics} (mean and standard deviation), hereafter referred to simply as ``statistics'', play a crucial role in model adaptation, aligning with recent observations \cite{schneider2020betterinc}.
However, the best choice of statistics, \eg, those of the source domain, test image, or a mix between those, is highly variable between different domain shifts, due to the instability of relying on only a single target domain image.
To address these challenges, we propose a novel method for creating an ensemble of several possible adapted models constructed using different estimates of the target domain statistics. Rather than simply selecting the model with the lowest prediction entropy, we integrate all models' predictions. 
We explore various integration strategies, including simple averaging, weighted averaging based on entropy or entropy sharpness (a concept recently discovered by \cite{niu2023sharpness} to be informative during TTA), and others.
This ensembling approach is robust to relying on only a single test domain image because it does not require iterative optimization of model parameters.
We also incorporate a novel approach of equally balancing the entropy contributions of predicted foreground and background pixels that is specifically designed for segmentation, rather than treating all pixel predictions equally.
Our method is named \textbf{\methodname}: \textbf{In}tegrated \textbf{T}est-time \textbf{Ent}ropy Weighting for Single Image Adaptation, summarized in Fig. \ref{fig:summary}.
InTEnt achieves superior performance over existing methods in a variety of medical image domain shift settings, as shown in Table \ref{tab:overall}.

\paragraph{Contributions.} Our main contributions are the following:
\begin{enumerate}
    \item To the best of our knowledge, we propose the first single-image TTA method for medical image segmentation.
    \item We demonstrate the importance of batch normalization layer statistic selection for adapting models to a single test image, and use this to generate an ensemble of possible adapted models.
    \item To address the variability of the optimal batch norm. statistic choice for different domain shift settings, we propose a simple yet effective strategy of integrating the predictions of the different adapted models, weighted by their prediction entropy.
    \item Our method achieves an average performance of $71.6\%$ Dice similarity coefficient (DSC) for $24$ different domain shift settings across $3$ datasets, while other approaches give at most $68.7\%$ DSC.
\end{enumerate}

\section{Related Work}
\subsection{Single Image Test-Time Adaptation}
Test-time adaptation (TTA) aims at fine-tuning model parameters during test time, using only test data \cite{liang2023comprehensive}.
In this work, we broaden this application by considering the challenging case of only having a single test image, focusing on medical image segmentation where such a constraint is very realistic.

Certain previous works approach single image TTA by learning extra information during training \cite{Hu2021FullyTA, liu2022single}. 
For example, \cite{karani2021test} introduced a denoising autoencoder to correct test predictions;
\cite{valanarasu2023onthefly} proposed to pre-train a domain encoder that can simulate target image domain information;
\cite{gao2023back} learned a diffusion model that projects target domain images back to the source domain.
Although effective, these methods utilize auxiliary networks during training, removing the possibility of adapting arbitrary pre-trained models to target downstream tasks.
Another direction is to augment the test image before adaptation, which increases the robustness and correctness in estimating from a single test image \cite{Khurana2021SITASI, zhang2022memo, gao2023back}.
However, these methods are sensitive to the choice of augmentation function, and we find that they lead to sub-optimal performance for segmentation. 

\subsection*{Test-time Adaptation with Prediction Entropy}
\cite{Wang2021TentFT} first found that minimizing prediction entropy during test time can improve network performance.
We claim that the effectiveness of this approach relies on two aspects: (1) the model is well-calibrated and (2) the estimation of the test domain prediction entropy is accurate.
A model is well-calibrated if its predicted probabilities are representative of the true correctness likelihood \cite{guo2017calibration}, \ie, a predicted probability closer to 0 or 1 (lower entropy) should be more likely to be correct.
Thus, entropy minimization results directly in more accurate predictions.
However, it is infeasible to precisely calibrate the model without access to the training data.

The second aspect is that prediction entropy can only be reliably estimated when test images arrive in a batch and are from the same domain, allowing for a faithful estimation of the target domain statistics via entropy minimization \cite{niu2023sharpness, gan2023decorate, zhao2023on}.
To solve this problem in the single-image TTA setting, we propose to instead integrate predictions made with different target domain statistic estimates, as only using a single estimate found via entropy minimization leads to inconsistent results due to the difficulty of accurate estimation with just one image. 

\section{Method}

\subsection{Single Image Test-Time Adaptation for Segmentation}


In single image test-time adaptation (SITTA), we have a pre-trained model $f$ with parameters $\theta$, and a \textit{single} test image $x\in \mathbb{R}^{N_C\times H \times W}$ with $N_C$ channels, which has an unknown corresponding label $Y$ sampled from some test domain conditional probability distribution $p_\mathrm{test}(Y|x)$ \cite{wang2022continual}.
For binary segmentation, an image label $Y\in\{0,1\}^{H\times W}$ is a segmentation mask of $H\times W$ pixels, and $f$ outputs a predicted probability map $\hat{P}\in[0,1]^{H\times W}$. The goal of TTA is to determine the optimal model parameters $\theta$ that maximize the likelihood $p(y^{i,j}|x,\theta)$ of the model prediction for each pixel $(i,j)$, given the unknown pixel label sampled from $p_\mathrm{test}(y^{i,j}|x)$. 

Let $q_\theta(y^{i,j}|x)$ denote the model's prediction distribution for a given pixel, \ie, $f$'s predicted probability for that pixel to contain the object of interest. Maximizing the pixel likelihood is equivalent to minimizing the cross-entropy between the predicted and true distributions, $H(q_\theta, p_\mathrm{test}) = -\mathbb{E}_{y^{i,j}\sim q_{\theta}(y^{i,j}|x)}\ln p_\mathrm{test}(y^{i,j}|x)$ \cite{ho2019real}.
Writing $q_{\theta}(y^{i,j}|x)$ and $p_\mathrm{test}(y^{i,j}|x)$ as $q_{\theta}$ and $p_\mathrm{test}$ for brevity, the cross-entropy can be decomposed into
\begin{align}
\label{eq:crossent}
H(q_\theta, p_\mathrm{test}) &= -\mathbb{E}_{y^{i,j}\sim  q_\theta}\ln p_\mathrm{test} \nonumber \\
        &= -\mathbb{E}_{y^{i,j}\sim  q_\theta}[\ln p_\mathrm{test} - \ln  q_\theta + \ln  q_\theta] \nonumber\\ 
        &= -\mathbb{E}_{y^{i,j}\sim  q_\theta}\ln  q_\theta + \mathbb{E}_{y^{i,j}\sim  q_\theta} \ln\frac{ q_\theta}{p_\mathrm{test}}  \nonumber \\
        &= H[ q_\theta] + D_\mathrm{KL}[ q_\theta || p_\mathrm{test}],
\end{align}
where $H[q_\theta]=-\mathbb{E}_{y^{i,j}\sim q_{\theta}}\ln q_\theta$ is the entropy of the predictive distribution, and $D_\mathrm{KL}$ is the Kullback-Leibler Divergence between $q_\theta$ and $p_\mathrm{test}$. 

Without access to the target labels, it is impossible to evaluate the $p_\mathrm{test}=p_\mathrm{test}(y^{i,j}|x)$ term in $D_\mathrm{KL}$, so that minimizing the prediction entropy $H[ q_\theta]$ would be the only feasible option.
If we assume that the predictions for different pixels are independent \cite{vu2019advent}, the predictive and true distributions for the entire image mask $Y$ can be written as the product of individual pixel probabilities, as
$q_{\theta}(Y|x) = \prod_{i=1}^H \prod_{j=1}^W q_{\theta}(y^{i,j}|x)$ and 
$p_\mathrm{test}(Y|x) = \prod_{i=1}^H \prod_{j=1}^W p_\mathrm{test}(y^{i,j}|x)$,
and similar for the likelihood 
$p(Y|x,\theta)=\prod_{i=1}^H \prod_{j=1}^W p(y^{i,j}|x,\theta)$.
Then, minimizing the mask prediction entropy can be accomplished by minimizing the sum (or equivalently, the average) of pixel prediction entropies,
\begin{align}
\label{eq:tgtentropy}
H[q_\theta (Y|x)] &=-\mathbb{E}_{Y\sim q_\theta (Y|x)}\ln q_\theta (Y|x) \nonumber\\
    &=-\mathbb{E}_{Y\sim q_\theta (Y|x)}\sum_{i=1}^H \sum_{j=1}^W\ln q_{\theta}(y^{i,j}|x).
\end{align}

\paragraph{Foregound-Background-Balanced Entropy Weighting}
Despite pixel predictions being independent, they can contribute differently to the final quality of the predicted mask.
For example, given one mask prediction with moderately low entropy across all pixels, and another with zero entropy for background predictions and high entropy for foreground predictions, the former would result in more faithful predictions yet a lower overall entropy if averaged across all pixels.
Thus, we propose a new strategy to balance the importance of foreground and background predictions.
Specifically, we define the predicted foreground entropy as
\begin{align}
\begin{split}
\label{eq:balanceentropy}
H_{FG}[q_\theta (Y|x)] &= -\mathbb{E}_{Y\sim q_\theta (Y|x)}\sum_{i,j \in S} \ln q_{\theta}(y^{i,j}|x), \\
\text{ where }S &= \{(i,j) \mid q_{\theta}(y^{i,j}|x) \ge 0.5 \},
\end{split}
\end{align}
with the background entropy $H_{BG}[q_\theta (Y|x)]$ defined similarly with the complement of $S$. We then use the average of $H_{FG}$ and $H_{BG}$ as the final weight for a given model prediction. 

\subsection{Adapting Models via Batch Normalization Layers}
\label{sec:BN} 
Formally, a Batch Normalization (BN) layer \cite{Ioffe2015BatchNA} can be expressed as
\begin{equation}
BN(h) = \gamma (h-\mu_{train})/\sigma_{train} + \beta,
\end{equation}
where $h$ is the input feature map, $\{\gamma, \beta\}$ are scale and shift parameters learned during training, and
$S_{train}:= \{ \mu_{train}, \sigma_{train} \}$ are the tracked mean and variance of the source domain.
When domain shift occurs, the test domain statistics $S_{test} :=\{\mu_{test}, \sigma_{test} \}$ can differ from the tracked ones, leading to suboptimal performance.
While other methods optimize ${\gamma, \beta}$ at test time with gradient descent to minimize prediction entropy \cite{Wang2021TentFT, niu2023sharpness} or customized objectives \cite{Hu2021FullyTA}, we find that in the SITTA setting, optimization leads to minor changes in the final prediction.
Therefore, we propose to instead modify the \textit{statistics}, with the following scheme.

We can freely interpolate between the training and test statistics $S_{train}$ and $S_{test}$ with $\lambda \in (0,1)$ to obtain mixed statistics
\begin{equation}
\label{eq:mix}
    S_{mix}^{\lambda} := \lambda \times S_{train} + (1-\lambda) \times S_{test}.
\end{equation}
Instead of selecting a single $\lambda$, we sample evenly from $(0, 1)$ with a step size hyperparameter $C$ to create a range of mixed statistics to consider. By default, we use $C=0.2$, which creates $\lambda \in \{0.2,0.4,0.6,0.8\}$.
We use each of the training, test, and different mixed statistics to define an ensemble of adapted models. Fig. \ref{fig:summary}c) visualizes how varying these statistics will affect model predictions, and we include a more detailed visualization in the Experiment section. 

\subsection{Integrating Over Adapted Models}

\label{sec:networkselect}

With our proposed strategy of adapting $f$ to the test domain via the modification of batch norm. statistics, we can obtain multiple predictions for a test image $x$ by using each of the statistics 
\begin{equation}
\label{eq:BNensemble}
S_k \in \{S_{train}, S_{mix}^{\lambda=C,...,1-C}, S_{test} \}
\end{equation}
to define a set $\mathcal{F}$ of models. A simple solution would be to use the model $f_k$ out of $\mathcal{F}$ that results in the prediction with minimum entropy, but we found this to be less stable and robust, due to relying on a single image from the target domain for entropy estimation.

Instead, we take a Bayesian approach \cite{berger1999integrated,hoeting1999bayesian} of integrating over (the predictions of) all adapted models, weighted by their likelihoods, to obtain an optimal prediction 
$\hat{P} = \int_{f_k\in\mathcal{F}} \hat{P}_{k} p(Y | x, \theta_k) d\theta_k$,
where $\hat{P}_{k}$ is the segmentation probability map prediction of model $f_k\in\mathcal{F}$ with adapted parameters $\theta_k$ (\textbf{note:} here we write $\theta_k$ to include the adapted batch norm. layer statistics, although these aren't learnable).
This model-averaging scheme is computationally tractable over our set $\mathcal{F}$ of finite models, giving 
\begin{equation}
\label{eq:modelsum}
\hat{P} = \sum_{f_k\in\mathcal{F}}  \hat{P}_{k} p(Y | x, \theta_k).
\end{equation}
As we cannot fully evaluate the likelihood of a model without the ground truth label for $x$, we can approximate it using the prediction entropy as in Eq. \eqref{eq:crossent}, with
\begin{align}
\label{eq:weightapprox}
    p(Y | x, \theta_k) &=  e^{-H(q_{\theta_k}, p_\mathrm{test})} \nonumber\\
    &= e^{-H(q_{\theta_k})}e^{- D_\mathrm{KL}(q_{\theta_k} || p_\mathrm{test})} \underset{\sim}{\propto} e^{-H(q_{\theta_k})}, 
\end{align}
where we have written $q_{\theta_k}$ and $p_\mathrm{test}$ short-hand for the predictive and true segmentation distributions $q_{\theta_k}(Y|x)$ and $p_\mathrm{test}(Y|x)$, respectively. In other words, models that have lower balanced segmentation prediction entropy: $w_k :=-H_{FG}[q_{\theta_k} (Y|x)]+H_{BG}[q_{\theta_k} (Y|x)]$ (Eq. \eqref{eq:balanceentropy}) are weighted higher.
Lastly, we normalize $w_k$ by $w_k' = w_k / [\max (\{w_k\}_{\forall k}) - \min (\{w_k\}_{\forall k})]$ to assign higher weights to predictions with lower entropy.
This is the integration strategy that we use for our final algorithm (performance shown in Table \ref{tab:overall}). We also compare a wide range of entropy-based prediction weighting strategies in Sec. \ref{sec:ablation}. We will next introduce entropy sharpness, a recent concept that is also potentially usable as a weighting strategy.

\begin{algorithm}[t]
    \caption{\textbf{In}tegrated \textbf{T}est-time \textbf{Ent}ropy Weighting for Single Image Adaptation for Segmentation}
    \label{alg:main}
    \textbf{Input}: Test image $x\in \mathbb{R}^{N_C\times H \times W}$, source domain-trained segmentation model $f:\mathbb{R}^{N_C\times H \times W}\rightarrow [0, 1]^{H\times W}$. \\
\begin{algorithmic}[1] 
    \STATE \textit{Create ensemble of adapted models by modifying batch norm statistics:}
    \STATE $\mathcal{F} = \{f_k : f \text{ with batch norm. stats $S_k$ \textit{(Eq. \eqref{eq:BNensemble})}}\}$
    \STATE \textit{Predict segmentation probability maps:} $\hat{P}_{k} = f_k(x)$
    \STATE \textit{Weight each model by its prediction entropy (Eq. \eqref{eq:balanceentropy}):}
    \STATE $w_k = -H_{FG}[q_{\theta_k}(Y|x)]-H_{BG}[q_{\theta_k}(Y|x)]$ 
    \STATE \textit{Normalize weights:}
    \STATE $w_k' = w_k / [\max (\{w_k\}_{\forall k}) - \min (\{w_k\}_{\forall k})]$
    \STATE $\{w_k\}_{\forall k}=\mathrm{softmax}(\{w_k'\}_{\forall k})$
    \STATE \textit{Obtain integrated segmentation prediction:}
    \STATE $\hat{P} = \sum_{f_k\in\mathcal{F}}  w_k\hat{P}_{k}$
\end{algorithmic}
\end{algorithm}

\subsection{Minimizing Prediction Entropy Sharpness}
As recent TTA literature \cite{niu2023sharpness} found that prediction entropy $H[q_{\theta_k}(Y|x)]$ can be unstable when estimated from a small number of test images, we also evaluate an alternative strategy to weight models according to prediction entropy {\it sharpness} with respect to model parameters.

The \textit{sharpness} of the prediction entropy of a model is defined as the entropy's highest possible sensitivity with respect to a small perturbation $\epsilon$ to the model parameters. Formally, finding model parameters that give minimum entropy sharpness is a joint optimization problem
\begin{equation}
\label{eq:sharpnessopt}
\min_\theta\max_{|| \epsilon||_2 \le \rho} H[q_{\theta_k + \epsilon}(Y|x)]
\end{equation}
\cite{niu2023sharpness}, for some small constant $\rho$ ($0.1$ by default), where $H[q_{\theta_k + \epsilon}(Y|x)]$ is the prediction entropy of the model evaluated with parameters $\theta_k + \epsilon$ on the test image $x$ (Eq. \eqref{eq:tgtentropy}). If a first-order Taylor approximation is used for the inner optimization, a closed-form solution
\begin{equation}
    \hat{\epsilon}(\theta)=\frac{\rho \operatorname{sign}\left(\nabla_{\theta} H[q_{\theta}(Y|x)] \right)\left|\nabla_{\theta} H[q_{\theta}(Y|x)] \right|} {\left\|\nabla_{\theta} H[q_{\theta}(Y|x)] \right\|_2}
\end{equation}
is possible \cite{foret2020sharpness}.
We can then easily estimate the prediction entropy sharpness of some adapted model $f_k\in\mathcal{F}$ as 
\begin{equation}
\label{eq:sharpness-general}
\mathrm{sharp}(f_k;x) = H[q_{\theta_k + \hat{\epsilon}(\theta_k)}(Y|x)] - H[q_{\theta_k}(Y|x)].
\end{equation}

Returning to our model-averaging scheme of the previous section, we can give high weight $w_k$ to the prediction of an adapted model $f_k$ if it has low entropy sharpness, to obtain a final integrated prediction $\hat{P} = \sum_{f_k\in\mathcal{F}}  w_k\hat{P}_{k}$. For our case of single image TTA for segmentation, the sharpness (Eq. \eqref{eq:sharpness-general}) simplifies to
\begin{align}
    \label{eq:sharpness}
    \mathrm{sharp}(f_k;x) &= \nonumber\\
    \sum_{i=1}^H \sum_{j=1}^W \hat{P}_{k}^{i,j} \ln \hat{P}_{k}^{i,j} - \hat{P}_{\theta_k + \hat{\epsilon}(\theta_k)}^{i,j} &\ln \hat{P}_{\theta_k + \hat{\epsilon}(\theta_k)}^{i,j}
\end{align}
(using Eq. \eqref{eq:tgtentropy}), where $\hat{P}_{\theta_k + \hat{\epsilon}(\theta_k)}^{i,j}$ is the $(i,j)$ entry of the predicted segmentation probability map of $f_k$ for $x$ given parameters $\theta_k + \hat{\epsilon}(\theta_k)$. We could then define model weights as $w_k := -\mathrm{sharp}(f_k;x)$, which we will later compare to our strategy. 



\begin{table*}[ht!]
\caption{\textbf{The performance of UNet with various TTA methods given different batch norm. layer statistic choices defined by $\lambda$}, given as Dice segmentation similarity score averaged over 10 repeated experiments. Models are tested on all target domains from the same dataset. The highest score in each choice is highlighted.}
\vskip 0.15in
\begin{center}
\begin{small}
\begin{sc}
\begin{tabular}{ll|cccc|ccc|ccc|c}
    \hline
    \multirow{2}{*}{$\lambda$} & \multirow{2}{*}{Method} & \multicolumn{4}{c|}{Spinal Cord} & \multicolumn{3}{c|}{Chest} & \multicolumn{3}{c|}{Retinal} & \multirow{2}{*}{Avg.} \\ \cline{3-12}
    & & Site1 & Site2 & Site3 & Site4 & CHN & MCU & JSRT & CHASE & HRF & RITE &\\
    \hline
    \rowcolor{Gray}
    \multirow{5}{*}{$1.0$} 
    & UNet     & $49.0$ & $72.9$ & $34.0$ & $75.3$ & $90.7$ & $\mathbf{80.4}$ & $77.5$ & $\mathbf{46.4}$ & $57.7$ & $55.9$ & $\mathbf{64.0}$\\
    & +Tent    & $48.2$ & $72.7$ & $\mathbf{34.3}$ & $75.9$ & $90.7$ & $80.1$ & $76.8$ & $45.9$ & $57.1$ & $55.3$ & $63.7$\\
    & +SAR     & $\mathbf{49.8}$ & $\mathbf{73.5}$ & $32.7$ & $74.8$ & $90.3$ & $80.3$ & $\mathbf{79.3}$ & $46.8$ & $\mathbf{58.2}$ & $\mathbf{56.3}$ & $64.2$\\
    & +FSeg    & $48.2$ & $72.8$ & $34.1$ & $\mathbf{75.9}$ & $\mathbf{90.7}$ & $80.1$ & $76.8$ & $45.8$ & $57.1$ & $55.3$ & $63.7$\\
    & +MEMO    & $47.7$ & $72.5$ & $33.8$ & $75.4$ & $90.0$ & $80.3$ & $75.8$ & $45.8$ & $57.0$ & $55.3$ & $63.4$\\
    \hline
    \rowcolor{Gray}
    \multirow{6}{*}{$0.5$} 
    & UNet     & $62.2$ & $70.4$ & $43.8$ & $77.3$ & $95.7$ & $91.5$ & $93.2$ & $54.8$ & $59.5$ & $61.5$ & $71.0$\\
    & +Tent    & $61.9$ & $70.3$ & $\mathbf{44.4}$ & $78.4$ & $95.7$ & $91.6$ & $93.4$ & $54.6$ & $59.1$ & $61.6$ & $71.1$\\
    & +SAR     & $62.3$ & $71.1$ & $41.5$ & $76.3$ & $95.5$ & $90.9$ & $92.9$ & $54.9$ & $59.6$ & $61.5$ & $70.7$\\
    & +FSeg    & $61.9$ & $70.3$ & $44.3$ & $78.4$ & $\mathbf{95.7}$ & $\mathbf{91.7}$ & $\mathbf{93.4}$ & $54.5$ & $59.1$ & $61.5$ & $71.1$\\
    & +MEMO    & $61.6$ & $70.0$ & $44.0$ & $78.4$ & $95.6$ & $91.3$ & $93.4$ & $54.5$ & $59.0$ & $61.5$ & $70.9$\\
    & +SITA    & $\mathbf{63.4}$ & $\mathbf{71.4}$ & $41.6$ & $78.1$ & $95.8$ & $91.3$ & $93.2$ & $\mathbf{55.4}$ & $\mathbf{60.2}$ & $\mathbf{61.6}$ & $\mathbf{71.2}$\\
    \hline 
    \rowcolor{Gray}
    \multirow{6}{*}{$0.0$} 
    & UNet     & $56.0$ & $65.8$ & $47.3$ & $63.2$ & $\mathbf{95.8}$ & $93.4$ & $89.8$ & $\mathbf{57.5}$ & $\mathbf{58.3}$ & $61.0$ & $68.8$\\
    & +Tent    & $53.2$ & $65.7$ & $47.5$ & $64.5$ & $95.3$ & $93.5$ & $90.3$ & $57.4$ & $58.2$ & $61.3$ & $68.7$\\
    & +SAR     & $54.9$ & $66.2$ & $47.0$ & $61.6$ & $95.1$ & $93.1$ & $89.3$ & $57.5$ & $58.3$ & $60.7$ & $68.4$\\
    & +FSeg    & $53.3$ & $65.7$ & $47.5$ & $64.5$ & $95.3$ & $\mathbf{93.5}$ & $90.3$ & $57.5$ & $58.2$ & $\mathbf{61.3}$ & $68.7$\\
    & +MEMO    & $53.8$ & $65.4$ & $\mathbf{47.8}$ & $64.5$ & $95.2$ & $93.3$ & $90.4$ & $57.3$ & $58.2$ & $\mathbf{61.3}$ & $68.7$\\
    & +SITA    & $\mathbf{56.0}$ & $\mathbf{67.1}$ & $44.5$ & $\mathbf{67.6}$ & $95.4$ & $93.2$ & $\mathbf{90.4}$ & $57.8$ & $58.2$ & $60.8$ & $\mathbf{69.1}$\\

    \hline 
\end{tabular}
\end{sc}
\end{small}
\end{center}
\vskip -0.1in
\label{tab:diff}
\end{table*}

\subsection{Summary}

We summarize our method for single image test-time adaptation for segmentation in Fig. \ref{fig:summary} and Algorithm \ref{alg:main}. Beginning with some segmentation model trained on source domain data and a single test image of an unknown domain that we wish to adapt the model to, we first use our batch norm. statistic modification scheme (Eq. \eqref{eq:BNensemble}) to create an ensemble of possible adapted models. By default, we weigh each model according to its segmentation prediction entropy before integrating over all models to obtain a final prediction, but we will also evaluate additional weighting strategies, including via entropy sharpness. We name our method \textbf{\methodname}, or \textbf{In}tegrated \textbf{T}est-time \textbf{Ent}ropy Weighting for Single Image Adaptation.

Our method takes about 0.06 seconds to compute for a single test image, consisting of 6 times forward (Eq. \eqref{eq:BNensemble} with $C=0.2$) where a single forward takes $0.01$ second on an NVIDIA RTX A6000.
Note that computation cost is not our primary concern given that we only have one image to perform inference on in the single-image TTA setting. 

\begin{figure}[t]
    \centering
    \includegraphics[width=\linewidth]{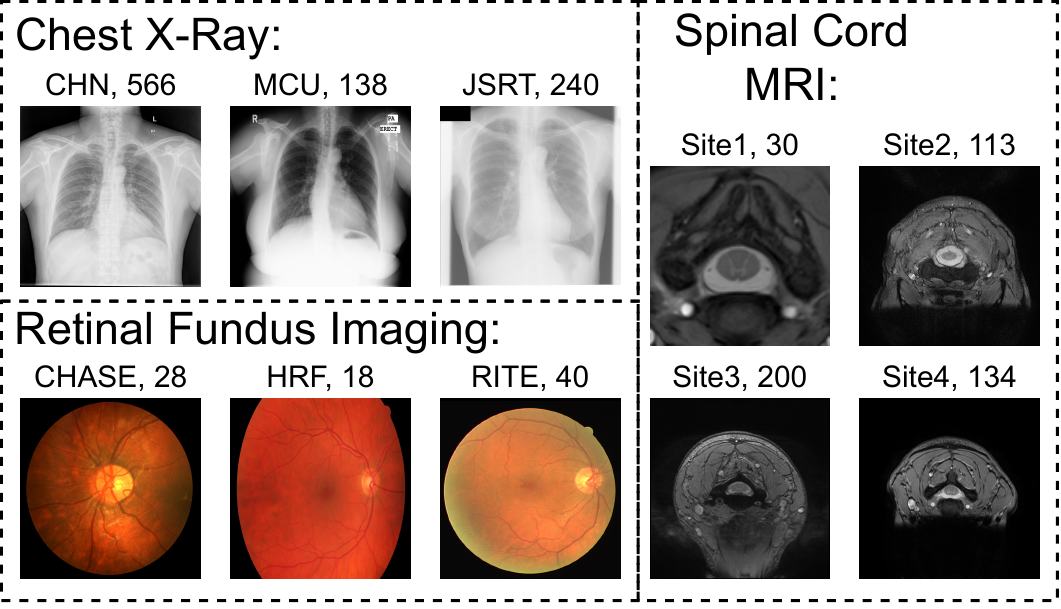}
    \caption{\textbf{Overview of the datasets used in this paper.} Above each example image, we list its domain and the total number of images from this domain.}
    \label{fig:dataset}
\end{figure}

\begin{figure*}[t] 
\vskip 0.2in
\centering
\includegraphics[width=\linewidth]{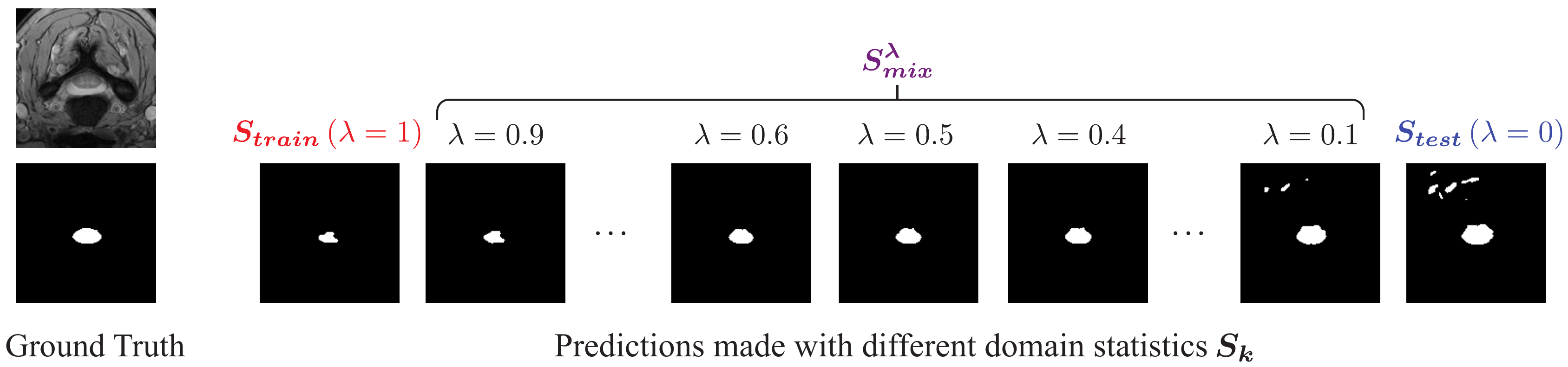}
\caption{The effect on model prediction when using different domain batch norm. layer statistics.}
\label{fig:stats}
\vskip -0.2in
\end{figure*}

\section{Experiments and Results}
\subsection{Setup}
\textbf{Datasets.} We evaluate our proposed method on three medical image segmentation tasks with publicly available multi-institution/domain datasets.
Grouped by \{modality\}/\{object of interest\}, these are: (1) \textbf{Spinal Cord (SC) MRI slices}/gray matter: Spinal Cord Gray Matter Segmentation Challenge Dataset  \cite{Prados2017SpinalCG}; (2) \textbf{Retinal (RET.) Fundus Imaging}/blood vessel: CHASE \cite{fraz2012ensemble}, RITE \cite{hu2013automated}, and HRF \cite{odstrvcilik2009improvement}; (3) \textbf{Chest (CHE.) X-ray}/lung: CHN, MCU \cite{Jaeger2014TwoPC} and JSRT \cite{Shiraishi2000DevelopmentOA}.
Figure \ref{fig:dataset} summarizes the domains in each modality and the number of images from each domain.
To evaluate various domain adaptation methods in a given modality, we train a segmentation model on images from a single domain and adapt and evaluate the model for one of the other domains. Model parameters are reset to their source domain setting following adaptation evaluation, as we consider the offline setting.

\textbf{Implementation Details.} We center crop the input images to $144\times 144$ for Spinal Cord \cite{li2020domain}, resize input images to $256\times 256$ for Fundus, and $128\times 128$ for Chest, following prior works.
All images are further normalized to $[0, 1]$.
We use an improved version of the UNet architecture \cite{Nichol2021ImprovedDD} for the segmentation model, which includes additional attention layers and a middle block between the encoder and decoder. The model is trained with equally weighted binary cross entropy (BCE) and Dice coefficient losses,
optimized using Adam \cite{kingmaAdam} with
a learning rate of $10^{-4}$ and momentum of $0.9$.
Batch size is set to $10$.
During training, the batch norm. layer statistics are updated via an exponentially moving average with a step size of $0.1$. Segmentation predictions are evaluated with the Dice similarity score with respect to the target mask. 
$80\%$ of the images are randomly selected for training and the rest is used for validation.
We train for $200$ epochs, with early-stopping criteria for when the (source domain) validation score is not improved after $20$ epochs. 
All experiments are repeated $10$ times with the same train/validation split. The average performance is reported. Code and trained models will be made publicly available upon acceptance.

\textbf{Competing methods.} We compare our method to several recent TTA approaches that can be extended to the SITTA setting. 
Tent \cite{Wang2021TentFT}, SAR \cite{niu2023sharpness}, and FSeg \cite{Hu2021FullyTA} propose to minimize entropy, entropy sharpness, and Regional Nuclear-Norm loss, respectively, by updating normalization layer parameters, which all use test image batch norm. statistics ($\lambda=0$ in Eq. \eqref{eq:mix}).
SITA \cite{Khurana2021SITASI} is another TTA strategy that takes the batch norm. statistics of different augmentations of the test image, and uses the average of all statistics to make the final prediction, using $\lambda=0.8$.
We also evaluate SITA with their additional proposed strategy ``OP'' for finding the optimal statistics interpolated between the train and test domains using majority voting on minimum entropy.
Finally, MEMO \cite{zhang2022memo} combines both strategies by computing the average prediction entropy given a set of transformed versions of the test image, using $\lambda={15}/{16}$.
To adapt these methods to the (offline) SITTA setting, we reduce the test batch size to 1 and reset the model parameters after each adaptation. All other hyperparameters follow the settings of the respective original paper.


\begin{table*}[t]
\caption{
\textbf{Top:} Baseline UNet performance with different batch norm. statistics $S_{mix}^{\lambda}$ (Eq. \eqref{eq:mix}), averaged over all domain shifts. \textbf{Bottom:} Integrated performance of the top block using different integration strategies.}
\vskip 0.15in
\begin{center}
\begin{small}
\begin{sc}
\begin{tabular}{ll|cccc|ccc|ccc|c}
    \hline
    \multirow{2}{*}{Method} & BN. stat. & \multicolumn{4}{c|}{Spinal Cord} & \multicolumn{3}{c|}{Chest} & \multicolumn{3}{c|}{Retinal} & \multirow{2}{*}{Avg.} \\ \cline{3-12}
    & Strategy & Site1 & Site2 & Site3 & Site4 & CHN & MCU & JSRT & CHASE & HRF & RITE &\\
    \hline
    \multirow{5}{*}{UNet} 
    & $\lambda=1.0$ & $49.0$ & $72.9$ & $34.0$ & $75.3$ & $90.7$ & $80.4$ & $77.5$ & $46.4$ & $57.7$ & $55.9$ & $64.0$\\
    & $\lambda=0.8$ & $57.8$ & $72.4$ & $37.7$ & $77.3$ & $94.1$ & $86.0$ & $91.1$ & $50.9$ & $59.0$ & $59.5$ & $68.6$\\
    & $\lambda=0.6$ & $61.6$ & $71.1$ & $41.6$ & $77.8$ & $95.4$ & $90.1$ & $93.3$ & $53.8$ & $59.4$ & $61.2$ & $70.5$\\
    & $\lambda=0.4$ & $62.0$ & $69.7$ & $45.8$ & $76.4$ & $95.8$ & $92.4$ & $92.8$ & $55.7$ & $59.5$ & $61.7$ & $71.2$\\
    & $\lambda=0.2$ & $59.7$ & $68.0$ & $49.6$ & $72.3$ & $95.7$ & $93.5$ & $91.6$ & $56.9$ & $59.1$ & $61.6$ & $70.8$\\
    & $\lambda=0.0$ & $56.0$ & $65.8$ & $47.3$ & $63.2$ & $95.8$ & $93.4$ & $89.8$ & $57.5$ & $58.3$ & $61.0$ & $68.8$\\
    \hline 
    \hline
    \multirow{8}{*}{InTEnt} 
    & Average    & $61.4$ & $70.5$ & $46.1$ & $78.7$ & $95.6$ & $91.7$ & $93.3$ & $54.7$ & $59.9$ & $61.8$ & $71.4$ \\
    & Entropy    & $61.4$ & $70.5$ & $46.6$ & $78.8$ & $95.6$ & $91.9$ & $93.4$ & $54.7$ & $59.9$ & $61.7$ & $71.5$ \\
    & Ent.-Min   & $52.2$ & $69.4$ & $47.6$ & $76.5$ & $95.0$ & $\mathbf{93.5}$ & $92.2$ & $53.2$ & $58.1$ & $56.3$ & $69.4$ \\
    & Ent.-Topk  & $59.6$ & $70.1$ & $\mathbf{49.5}$ & $78.6$ & $95.7$ & $93.4$ & $93.5$ & $54.4$ & $59.0$ & $59.5$ & $71.3$ \\
    & Ent.-Norm  & $60.7$ & $70.2$ & $49.2$ & $\mathbf{79.2}$ & $95.8$ & $92.9$ & $93.7$ & $54.7$ & $59.7$ & $61.0$ & $71.7$ \\
    & Ent.-Baln  & $\mathbf{62.2}$ & $\mathbf{71.4}$ & $45.2$ & $79.2$ & $\mathbf{95.9}$ & $92.9$ & $\mathbf{94.1}$ & $53.8$ & $\mathbf{59.9}$ & $\mathbf{62.2}$ & $71.7$ \\
    & Sharpness & $61.1$ & $70.5$ & $46.8$ & $77.6$ & $95.6$ & $92.3$ & $93.3$ & $\mathbf{55.7}$ & $59.7$ & $61.7$ & $71.4$ \\
    
\end{tabular}
\end{sc}
\end{small}
\end{center}
\vskip -0.1in
\label{tab:internalavged}
\end{table*}

\subsection{Performance Comparison to Existing Methods}
\label{sec:exp}

Table \ref{tab:overall} shows the average performance of existing TTA methods and our method across different datasets, averaged over all source/target domain splits. 
Our method achieves the leading performance on both the spinal cord and chest dataset, surpassing the runner-up methods by $3.2\%$ and $1.0\%$ Dice similarity score (DSC), respectively.
The method is also on par with SOTA on the fundus dataset ($0.3\%$ lower DSC).
Interestingly, we observe that the performance of Tent \cite{Wang2021TentFT}, SAR \cite{niu2023sharpness}, and FSeg \cite{Hu2021FullyTA} are quite similar despite their different optimization objectives; we will study this further as follows.

\subsection{Importance of the Choice of Batch Norm. Statistics}
The aforementioned adaptation methods (Tent, SAR, and FSeg) all utilize the test image batch norm. (BN) statistics, \ie, $\lambda=0$ (Eq. \eqref{eq:mix}).
To further study the relationship between BN statistics and adapted network performance, we evaluate all competing methods and the baseline model with different selections of statistics, \ie, choices of $\lambda$. 
Note that SITA is not applicable when $\lambda=1$ since it only alters the test statistics.
The results are shown in Tables \ref{tab:diff}, with the details of individual source/target domain performances in Appendix \ref{sec:detaildiff}. 

By comparing all methods for a fixed $\lambda$/statistic, we observe that the gains by each method are small in the single image segmentation TTA setting: usually $<1\%$ change in Dice similarity score (DSC), with the best being $+4.8\%$ DSC (Chest X-Ray, JSRT$\to$MCU, $\lambda=1$), and the worst being $-3.2\%$ DSC (SC, Site 1$\to$ Site 2, $\lambda=0$). 
The rank of these competing methods is also not consistent, especially when examined at the individual source/target domain level, further showing their instability.
Instead, the effect of the specific domain shift and statistics used is far greater.
Altering the hyperparameters (\eg, iteration count, learning rate) of the TTA methods could potentially amplify their effects, but this could also worsen cases where the method degraded performance. As such, we use the default hyperparameters recommended by each paper.

It could be the case that there is some optimal $\lambda$/mixture of source and test domain statistics for general single-image (segmentation) TTA, but our experiments do not support this. As shown in Table \ref{tab:diff}, the choice of optimal statistics can vary greatly even for different domain shifts within the same dataset, for both TTA-adapted models and UNet. For example, in the spinal cord dataset, when ``site 1'' becomes the source domain, a mix of train/test $(\lambda=0.5)$ is favored, yet when training on ``site 3'', models favor the test statistics $(\lambda=0.0)$. This was our motivation for instead \textit{integrating} over predictions made with a variety of statistics. We evaluate different integration/weighting strategies in the following section.

\subsection{Ablation Study: Adapted Model Integration Strategies}
\label{sec:ablation}
After creating an ensemble of adapted models $f_k$ using different statistics (Eq. \eqref{eq:BNensemble}), our default strategy for integrating over all models' predictions $\hat{P}_k$ is to weight each prediction by its balanced entropy between foreground and background, normalize the weights, and take a weighted average of the predictions (Algorithm \ref{alg:main}). 
We first present visually an example of how model predictions change to the change of $\lambda$/batch norm. statistic.
In Figure \ref{fig:stats}, we observe that the un-adapted model $(\lambda=1)$ fails to segment the gray matter fully and achieves an ideal prediction when the $\lambda\simeq0.5$, or about an even mix between train and test statistics. However, the model becomes over-confident and incorrectly segments the non-gray matter regions when the statistics come mainly from the given test image $(\lambda\simeq 0)$. 
Next, we evaluate a range of modifications to this strategy:
\begin{enumerate}
    \item ``Average'': average all predictions with equal weights.
    \item ``Entropy'': use the exact prediction entropy to weight, as $w_k =-H[q_{\theta_k}(Y|x)]$ (Eq. \eqref{eq:tgtentropy}).
    \item ``Ent.-Min'': the predictions with the minimum entropy are used as the final prediction.
    \item ``Ent.-TopK'': the top-k predictions with the minimum entropy as averaged. We set $K=2$. This is also the ``optimal prior'' (OP) method proposed in SITA. 
    \item ``Ent.-Norm'': a normalization is applied to ensure the maximum difference among the entropies is $1$.
    \item ``Ent.-Baln'': the entropy is computed separately for fore/background (Eq. \eqref{eq:balanceentropy}). We select this strategy when compared externally and the details are shown in Algorithm \ref{alg:main}.
    \item ``Sharpness'': use entropy sharpness (Eq. \eqref{eq:sharpness}) to weight, as $w_k = -\mathrm{sharp}(f_k; x)$, followed by the same weight normalization as in Algorithm \ref{alg:main}.
\end{enumerate}

We show the performance of our method using these different integration strategies, alongside the baseline model performance given the different batch norm. statistic / values of $\lambda$ being integrated over, in Table \ref{tab:internalavged}. The detailed performances of individual source/target domains are shown in Appendix \ref{sec:detailinternal}.

First, we observe that the ``Ent.-Min'' strategy usually leads to the worst performance among all weighting strategies, demonstrating the instability in relying on a single prediction/statistic.
To be noted, this strategy still gives an average performance of $69.4\%$ DSC, which is higher than the leading competing methods.
The novel concept of entropy sharpness (``Sharpness'') results in the best average performance on the Retinal dataset, yet this trend is not universal.
Although the ``Ent.-Baln'' strategy, which our method uses, surpasses other strategies in most scenarios, the difference in performance between the weighting strategies is small.
This may be caused by the variability of the relation between entropy and prediction correctness, where the root issue is trying to use a single data point to estimate prediction entropy.

In general, we see that the specific integration strategy used does not have a significant effect, including the choice of $C$ (Appendix \ref{sec:abla_fig}).
We argue that the main contribution of this work is to explore the importance and necessity of batch norm. statistic selection in TTA. Integrating predictions given multiple statistics is but one solution, and we hope our work can inspire further research in this direction.

\section{Conclusion}
Single-image test-time adaptation is attractive for medical image segmentation due to common imaging domain inhomogeneity issues, and the expense and difficulty of acquiring new target domain images. It also benefits the general TTA setting when applied to real-world scenarios. However, relying on only a single target domain image to perform adaptation comes with its difficulties and surprises; for example, using solely the test image batch norm. statistics is not always optimal.
Our proposed method, \textbf{\methodname}, stabilizes adapted model predictions by integrating over predictions made with multiple possible estimations of the target domain statistics.
We hope that our study motivates further research in segmentation SITTA for medical imaging and beyond, especially regarding the importance of the choice of normalization layer statistics.

\section*{Impact Statement}
This paper presents work whose goal is to advance the field of Machine Learning. There are many potential societal consequences of our work, none of which we feel must be specifically highlighted here.

\bibliography{main}

\begin{thebibliography}{38}
\providecommand{\natexlab}[1]{#1}
\providecommand{\url}[1]{\texttt{#1}}
\expandafter\ifx\csname urlstyle\endcsname\relax
  \providecommand{\doi}[1]{doi: #1}\else
  \providecommand{\doi}{doi: \begingroup \urlstyle{rm}\Url}\fi

\bibitem[Berger et~al.(1999)Berger, Liseo, and Wolpert]{berger1999integrated}
Berger, J.~O., Liseo, B., and Wolpert, R.~L.
\newblock Integrated likelihood methods for eliminating nuisance parameters.
\newblock \emph{Statistical science}, pp.\  1--22, 1999.

\bibitem[Foret et~al.(2020)Foret, Kleiner, Mobahi, and Neyshabur]{foret2020sharpness}
Foret, P., Kleiner, A., Mobahi, H., and Neyshabur, B.
\newblock Sharpness-aware minimization for efficiently improving generalization.
\newblock In \emph{International Conference on Learning Representations}, 2020.

\bibitem[Fraz et~al.(2012)Fraz, Remagnino, Hoppe, Uyyanonvara, Rudnicka, Owen, and Barman]{fraz2012ensemble}
Fraz, M.~M., Remagnino, P., Hoppe, A., Uyyanonvara, B., Rudnicka, A.~R., Owen, C.~G., and Barman, S.~A.
\newblock An ensemble classification-based approach applied to retinal blood vessel segmentation.
\newblock \emph{IEEE Transactions on Biomedical Engineering}, 59\penalty0 (9):\penalty0 2538--2548, 2012.

\bibitem[Gan et~al.(2023)Gan, Bai, Lou, Ma, Zhang, Shi, and Luo]{gan2023decorate}
Gan, Y., Bai, Y., Lou, Y., Ma, X., Zhang, R., Shi, N., and Luo, L.
\newblock Decorate the newcomers: Visual domain prompt for continual test time adaptation.
\newblock In \emph{Proceedings of the AAAI Conference on Artificial Intelligence}, volume~37, pp.\  7595--7603, 2023.

\bibitem[Gao et~al.(2023)Gao, Zhang, Liu, Darrell, Shelhamer, and Wang]{gao2023back}
Gao, J., Zhang, J., Liu, X., Darrell, T., Shelhamer, E., and Wang, D.
\newblock Back to the source: Diffusion-driven adaptation to test-time corruption.
\newblock In \emph{Proceedings of the IEEE/CVF Conference on Computer Vision and Pattern Recognition}, pp.\  11786--11796, 2023.

\bibitem[Guan \& Liu(2021)Guan and Liu]{guan2021domain}
Guan, H. and Liu, M.
\newblock Domain adaptation for medical image analysis: a survey.
\newblock \emph{IEEE Transactions on Biomedical Engineering}, 69\penalty0 (3):\penalty0 1173--1185, 2021.

\bibitem[Guo et~al.(2017)Guo, Pleiss, Sun, and Weinberger]{guo2017calibration}
Guo, C., Pleiss, G., Sun, Y., and Weinberger, K.~Q.
\newblock On calibration of modern neural networks.
\newblock In \emph{International conference on machine learning}, pp.\  1321--1330. PMLR, 2017.

\bibitem[Ho \& Wookey(2019)Ho and Wookey]{ho2019real}
Ho, Y. and Wookey, S.
\newblock The real-world-weight cross-entropy loss function: Modeling the costs of mislabeling.
\newblock \emph{IEEE access}, 8:\penalty0 4806--4813, 2019.

\bibitem[Hoeting et~al.(1999)Hoeting, Madigan, Raftery, and Volinsky]{hoeting1999bayesian}
Hoeting, J.~A., Madigan, D., Raftery, A.~E., and Volinsky, C.~T.
\newblock Bayesian model averaging: a tutorial (with comments by m. clyde, david draper and ei george, and a rejoinder by the authors.
\newblock \emph{Statistical science}, 14\penalty0 (4):\penalty0 382--417, 1999.

\bibitem[Hu et~al.(2021)Hu, Song, Gu, Luo, Chen, Chen, Zhang, and Zhang]{Hu2021FullyTA}
Hu, M., Song, T., Gu, Y., Luo, X., Chen, J., Chen, Y., Zhang, Y., and Zhang, S.
\newblock Fully test-time adaptation for image segmentation.
\newblock In \emph{International Conference on Medical Image Computing and Computer-Assisted Intervention}, 2021.

\bibitem[Hu et~al.(2013)Hu, Abr{\`a}moff, and Garvin]{hu2013automated}
Hu, Q., Abr{\`a}moff, M.~D., and Garvin, M.~K.
\newblock Automated separation of binary overlapping trees in low-contrast color retinal images.
\newblock In \emph{Medical Image Computing and Computer-Assisted Intervention--MICCAI 2013: 16th International Conference, Nagoya, Japan, September 22-26, 2013, Proceedings, Part II 16}, pp.\  436--443. Springer, 2013.

\bibitem[Ioffe \& Szegedy(2015)Ioffe and Szegedy]{Ioffe2015BatchNA}
Ioffe, S. and Szegedy, C.
\newblock Batch normalization: Accelerating deep network training by reducing internal covariate shift.
\newblock In \emph{International Conference on Machine Learning}, 2015.

\bibitem[Jaeger et~al.(2014)Jaeger, Candemir, Antani, Wang, Lu, and Thoma]{Jaeger2014TwoPC}
Jaeger, S., Candemir, S., Antani, S., Wang, Y.-X.~J., Lu, P., and Thoma, G.
\newblock Two public chest x-ray datasets for computer-aided screening of pulmonary diseases.
\newblock \emph{Quantitative imaging in medicine and surgery}, 4 6:\penalty0 475--7, 2014.

\bibitem[Karani et~al.(2021)Karani, Erdil, Chaitanya, and Konukoglu]{karani2021test}
Karani, N., Erdil, E., Chaitanya, K., and Konukoglu, E.
\newblock Test-time adaptable neural networks for robust medical image segmentation.
\newblock \emph{Medical Image Analysis}, 68:\penalty0 101907, 2021.

\bibitem[Khurana et~al.(2021)Khurana, Paul, Rai, Biswas, and Aggarwal]{Khurana2021SITASI}
Khurana, A., Paul, S., Rai, P., Biswas, S., and Aggarwal, G.
\newblock Sita: Single image test-time adaptation.
\newblock \emph{ArXiv}, abs/2112.02355, 2021.

\bibitem[Kingma \& Ba(2015)Kingma and Ba]{kingmaAdam}
Kingma, D.~P. and Ba, J.
\newblock Adam: {A} method for stochastic optimization.
\newblock In Bengio, Y. and LeCun, Y. (eds.), \emph{3rd International Conference on Learning Representations, {ICLR} 2015, San Diego, CA, USA, May 7-9, 2015, Conference Track Proceedings}, 2015.
\newblock URL \url{http://arxiv.org/abs/1412.6980}.

\bibitem[Koh et~al.(2020)Koh, Sagawa, Marklund, Xie, Zhang, Balsubramani, Hu, Yasunaga, Phillips, Beery, Leskovec, Kundaje, Pierson, Levine, Finn, and Liang]{Koh2020WILDSAB}
Koh, P.~W., Sagawa, S., Marklund, H., Xie, S.~M., Zhang, M., Balsubramani, A., Hu, W., Yasunaga, M., Phillips, R.~L., Beery, S., Leskovec, J., Kundaje, A., Pierson, E., Levine, S., Finn, C., and Liang, P.
\newblock Wilds: A benchmark of in-the-wild distribution shifts.
\newblock \emph{ArXiv}, abs/2012.07421, 2020.
\newblock URL \url{https://api.semanticscholar.org/CorpusID:229156320}.

\bibitem[Konz \& Mazurowski(2023)Konz and Mazurowski]{konz2023reverse}
Konz, N. and Mazurowski, M.~A.
\newblock Reverse engineering breast mris: Predicting acquisition parameters directly from images.
\newblock In \emph{Medical Imaging with Deep Learning}, 2023.

\bibitem[Li et~al.(2020)Li, Wang, Wan, Wang, Li, and Kot]{li2020domain}
Li, H., Wang, Y., Wan, R., Wang, S., Li, T.-Q., and Kot, A.
\newblock Domain generalization for medical imaging classification with linear-dependency regularization.
\newblock \emph{Advances in neural information processing systems}, 33:\penalty0 3118--3129, 2020.

\bibitem[Liang et~al.(2023)Liang, He, and Tan]{liang2023comprehensive}
Liang, J., He, R., and Tan, T.
\newblock A comprehensive survey on test-time adaptation under distribution shifts.
\newblock \emph{arXiv preprint arXiv:2303.15361}, 2023.

\bibitem[Liu et~al.(2018)Liu, Pan, Li, Chen, Tang, Lu, and Wang]{liu2018applications}
Liu, J., Pan, Y., Li, M., Chen, Z., Tang, L., Lu, C., and Wang, J.
\newblock Applications of deep learning to mri images: A survey.
\newblock \emph{Big Data Mining and Analytics}, 1\penalty0 (1):\penalty0 1--18, 2018.

\bibitem[Liu et~al.(2022)Liu, Chen, Dou, and Heng]{liu2022single}
Liu, Q., Chen, C., Dou, Q., and Heng, P.-A.
\newblock Single-domain generalization in medical image segmentation via test-time adaptation from shape dictionary.
\newblock In \emph{Proceedings of the AAAI Conference on Artificial Intelligence}, volume~36, pp.\  1756--1764, 2022.

\bibitem[Liu et~al.(2021)Liu, Xing, Yang, Fakhri, and Woo]{Liu2021AdaptingOS}
Liu, X., Xing, F., Yang, C., Fakhri, G.~E., and Woo, J.
\newblock Adapting off-the-shelf source segmenter for target medical image segmentation.
\newblock In \emph{Medical Image Computing and Computer Assisted Intervention -- MICCAI 2021}, pp.\  549--559, 2021.

\bibitem[Ma et~al.(2022)Ma, Chen, Zheng, Qin, Zhang, and Dou]{Ma2022TesttimeAW}
Ma, W., Chen, C., Zheng, S., Qin, J., Zhang, H., and Dou, Q.
\newblock Test-time adaptation with calibration of medical image classification nets for label distribution shift.
\newblock In \emph{International Conference on Medical Image Computing and Computer-Assisted Intervention}, 2022.

\bibitem[Nichol \& Dhariwal(2021)Nichol and Dhariwal]{Nichol2021ImprovedDD}
Nichol, A.~Q. and Dhariwal, P.
\newblock Improved denoising diffusion probabilistic models.
\newblock In \emph{International Conference on Machine Learning}, pp.\  8162--8171. PMLR, 2021.

\bibitem[Niu et~al.(2023)Niu, Wu, Zhang, Wen, Chen, Zhao, and Tan]{niu2023sharpness}
Niu, S., Wu, J., Zhang, Y., Wen, Z., Chen, Y., Zhao, P., and Tan, M.
\newblock Towards stable test-time adaptation in dynamic wild world.
\newblock In \emph{The Eleventh International Conference on Learning Representations}, 2023.

\bibitem[Odstr{\v{c}}il{\'\i}k et~al.(2009)Odstr{\v{c}}il{\'\i}k, Jan, Gaz{\'a}rek, and Kol{\'a}{\v{r}}]{odstrvcilik2009improvement}
Odstr{\v{c}}il{\'\i}k, J., Jan, J., Gaz{\'a}rek, J., and Kol{\'a}{\v{r}}, R.
\newblock Improvement of vessel segmentation by matched filtering in colour retinal images.
\newblock In \emph{World Congress on Medical Physics and Biomedical Engineering, September 7-12, 2009, Munich, Germany: Vol. 25/11 Biomedical Engineering for Audiology, Ophthalmology, Emergency \& Dental Medicine}, pp.\  327--330. Springer, 2009.

\bibitem[Prados et~al.(2017)Prados, Ashburner, Blaiotta, Brosch, Carballido-Gamio, Cardoso, Conrad, Datta, D{\'a}vid, Leener, Dupont, Freund, Wheeler-Kingshott, Grussu, Henry, Landman, Ljungberg, Lyttle, Ourselin, Papinutto, Saporito, Schlaeger, Smith, Summers, Tam, Yiannakas, Zhu, and Cohen-Adad]{Prados2017SpinalCG}
Prados, F., Ashburner, J., Blaiotta, C., Brosch, T., Carballido-Gamio, J., Cardoso, M., Conrad, B.~N., Datta, E., D{\'a}vid, G., Leener, B., Dupont, S., Freund, P., Wheeler-Kingshott, C., Grussu, F., Henry, R., Landman, B., Ljungberg, E., Lyttle, B., Ourselin, S., Papinutto, N., Saporito, S., Schlaeger, R., Smith, S.~A., Summers, P., Tam, R., Yiannakas, M., Zhu, A., and Cohen-Adad, J.
\newblock Spinal cord grey matter segmentation challenge.
\newblock \emph{Neuroimage}, 152:\penalty0 312 -- 329, 2017.

\bibitem[Quionero-Candela et~al.(2009)Quionero-Candela, Sugiyama, Schwaighofer, and Lawrence]{QuioneroCandela2009DatasetSI}
Quionero-Candela, J., Sugiyama, M., Schwaighofer, A., and Lawrence, N.~D.
\newblock Dataset shift in machine learning.
\newblock 2009.

\bibitem[Schneider et~al.(2020)Schneider, Rusak, Eck, Bringmann, Brendel, and Bethge]{schneider2020betterinc}
Schneider, S., Rusak, E., Eck, L., Bringmann, O., Brendel, W., and Bethge, M.
\newblock Improving robustness against common corruptions by covariate shift adaptation.
\newblock \emph{Advances in neural information processing systems}, 33:\penalty0 11539--11551, 2020.

\bibitem[Shiraishi et~al.(2000)Shiraishi, Katsuragawa, Ikezoe, Matsumoto, Kobayashi, Komatsu, Matsui, Fujita, Kodera, and Doi]{Shiraishi2000DevelopmentOA}
Shiraishi, J., Katsuragawa, S., Ikezoe, J., Matsumoto, T., Kobayashi, T., Komatsu, K., Matsui, M., Fujita, H., Kodera, Y., and Doi, K.
\newblock Development of a digital image database for chest radiographs with and without a lung nodule: receiver operating characteristic analysis of radiologists' detection of pulmonary nodules.
\newblock \emph{AJR. American journal of roentgenology}, 174 1:\penalty0 71--4, 2000.

\bibitem[Sun et~al.(2019)Sun, Wang, Liu, Miller, Efros, and Hardt]{Sun2019TestTimeTW}
Sun, Y., Wang, X., Liu, Z., Miller, J., Efros, A.~A., and Hardt, M.
\newblock Test-time training with self-supervision for generalization under distribution shifts.
\newblock In \emph{International Conference on Machine Learning}, 2019.

\bibitem[Valanarasu et~al.(2023)Valanarasu, Guo, VS, and Patel]{valanarasu2023onthefly}
Valanarasu, J. M.~J., Guo, P., VS, V., and Patel, V.~M.
\newblock On-the-fly test-time adaptation for medical image segmentation.
\newblock In \emph{Medical Imaging with Deep Learning}, 2023.
\newblock URL \url{https://openreview.net/forum?id=UQDalTzrEg}.

\bibitem[Vu et~al.(2019)Vu, Jain, Bucher, Cord, and P{\'e}rez]{vu2019advent}
Vu, T.-H., Jain, H., Bucher, M., Cord, M., and P{\'e}rez, P.
\newblock Advent: Adversarial entropy minimization for domain adaptation in semantic segmentation.
\newblock In \emph{Proceedings of the IEEE/CVF conference on computer vision and pattern recognition}, pp.\  2517--2526, 2019.

\bibitem[Wang et~al.(2021)Wang, Shelhamer, Liu, Olshausen, and Darrell]{Wang2021TentFT}
Wang, D., Shelhamer, E., Liu, S., Olshausen, B.~A., and Darrell, T.
\newblock Tent: Fully test-time adaptation by entropy minimization.
\newblock In \emph{International Conference on Learning Representations}, 2021.

\bibitem[Wang et~al.(2022)Wang, Fink, Van~Gool, and Dai]{wang2022continual}
Wang, Q., Fink, O., Van~Gool, L., and Dai, D.
\newblock Continual test-time domain adaptation.
\newblock In \emph{Proceedings of the IEEE/CVF Conference on Computer Vision and Pattern Recognition}, pp.\  7201--7211, 2022.

\bibitem[Zhang et~al.(2022)Zhang, Levine, and Finn]{zhang2022memo}
Zhang, M., Levine, S., and Finn, C.
\newblock Memo: Test time robustness via adaptation and augmentation.
\newblock \emph{Advances in Neural Information Processing Systems}, 35:\penalty0 38629--38642, 2022.

\bibitem[Zhao et~al.(2023)Zhao, Liu, Alahi, and Lin]{zhao2023on}
Zhao, H., Liu, Y., Alahi, A., and Lin, T.
\newblock On pitfalls of test-time adaptation.
\newblock In \emph{ICLR 2023 Workshop on Pitfalls of limited data and computation for Trustworthy ML}, 2023.
\newblock URL \url{https://openreview.net/forum?id=0Go_RsG_dYn}.

\end{thebibliography}
\bibliographystyle{icml2024}

\clearpage
\appendix

\onecolumn
\section{Detailed results}
\subsection{Baseline and other TTA approaches with different batch normalization layer statistics}
\label{sec:detaildiff}

Due to space limitation, we present all detailed results of Table \ref{tab:diff} (main paper) in Appendix Table \ref{tab:diff1}, \ref{tab:diff2}, and \ref{tab:diff3}. These tables show all results for each source/target split, rather than averaging over the domain shifts for each source domain as in Table \ref{tab:diff}. As we can observe, in general there are minor improvements or even degradation of other TTA methods given the instability in the single image TTA setting.

\subsection{Detailed results: baseline model performance with different batch norm statistics, and integration strategies}
\label{sec:detailinternal}
Due to space limitations, we present all detailed results of Table 6 (main paper) in Appendix Table \ref{tab:internal1}, \ref{tab:internal2}, and \ref{tab:internal3}. These tables show all results for each domain shift (columns), rather than averaging over the domain shifts for each source domain as in Table 6.
As can be observed in the top blocks of these tables, there is no universal optimal value of $\lambda$ used to determine the best batch norm. statistics to adapt the model, motivating our method to integrate over multiple $\lambda$.
The bottom blocks also suggest that the performance ranking of integration strategies varies between different domain shifts/train-test splits.
Our method's integration strategy, ``Norm'', gives the overall leading performance across the three datasets.

\section{Effects of hyperparameter choices}
\label{sec:abla_fig}
\begin{figure}[!h]
    \centering
    \includegraphics[width=\linewidth]{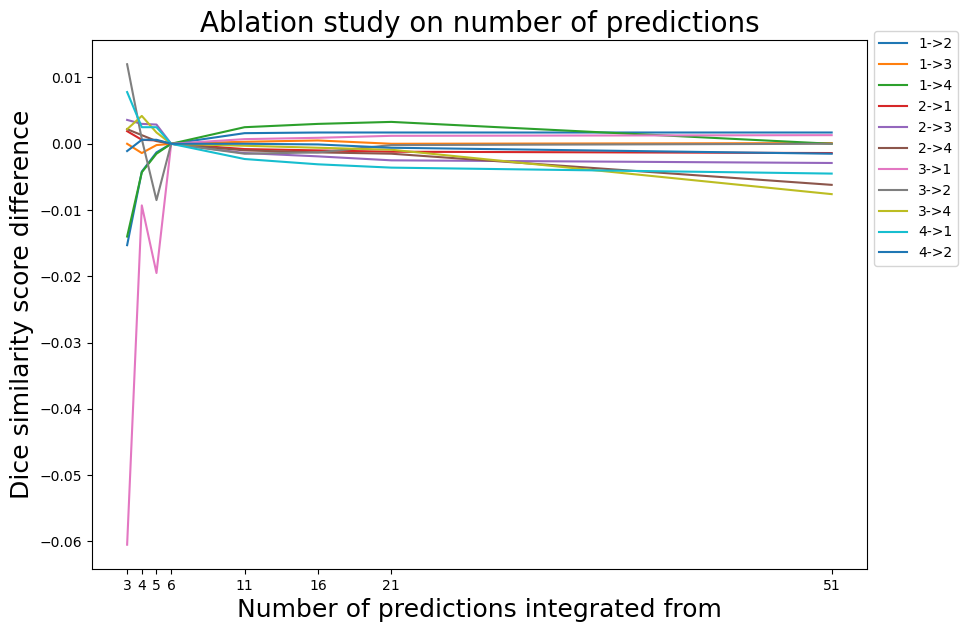}
    \caption{Our method's performance dependence on the choice of ensembled adapted model count (Eq. 5, main paper) for the SC dataset, over difference domain shifts.}
    \label{fig:enter-label}
\end{figure}

Our method's performance dependence on the choice of ensembled adapted model count (governed by $C$) is provided in Fig. \ref{fig:enter-label} for the SC dataset, over difference domain shifts. We report the change of Dice similarity score to highlight the affect of $C$. We observe no significant difference over a range of values from $2$ $(C=1)$ to $50$ $(C\simeq 0.02)$ models. Our model uses $C=0.2$ by default, or 6 models total.

\begin{table*}[h]
\fontsize{9pt}{9pt}\selectfont
\caption{
The performance (segmentation Dice coeff., avg. of $10$ repeated experiments) of other TTA approaches for different domain/site shifts using different batch norm. layer statistic choices $(\lambda)$, {\bf on the Spinal Cord (SC) dataset.} The best performance for a given $\lambda$ and domain shift is bolded.
The statistics used largely determine performance, while TTA methods themselves affect little in the performance, and may have identical values due to roundup.
}
\vskip 0.15in
\begin{center}
\begin{small}
\begin{sc}
\begin{tabular}{ll|cccccccccccc|c}
    \hline
    $\lambda$ & Method & $1\!\to\!2$ & $1\!\to\!3$ & $1\!\to\!4$ & $2\!\to\!1$ & $2\!\to\!3$ & $2\!\to\!4$ & $3\!\to\!1$ & $3\!\to\!2$ & $3\!\to\!4$ & $4\!\to\!1$ & $4\!\to\!2$ & $4\!\to\!3$ & Avg.$\uparrow$\\
    \hline
    \rowcolor{Gray}
    \multirow{5}{*}{$1.0$} & UNet & $71.2$ & $17.4$ & $58.5$ & $97.2$ & $39.4$ & $82.2$ & $26.2$ & $1.3$ & $74.5$ & $86.0$ & $67.2$ & $72.7$ & $\mathbf{57.8}$\\ 
    &+Tent  & $70.5$ & $16.8$ & $57.4$ & $\mathbf{97.3}$ & $38.9$ & $82.1$ & $\mathbf{26.3}$ & $\mathbf{1.4}$ & $\mathbf{75.3}$ & $\mathbf{87.0}$ & $\mathbf{67.9}$ & $\mathbf{72.9}$ & $\mathbf{57.8}$\\ 
    &+SAR   & $\mathbf{72.1}$ & $\mathbf{17.5}$ & $\mathbf{59.9}$ & $97.2$ & $\mathbf{40.8}$ & $\mathbf{82.5}$ & $26.2$ & $1.3$ & $70.6$ & $85.1$ & $66.6$ & $72.7$ & $57.7$\\ 
    &+FSeg  & $70.5$ & $16.9$ & $57.4$ & $\mathbf{97.3}$ & $39.0$ & $82.1$ & $25.9$ & $\mathbf{1.4}$ & $75.1$ & $\mathbf{87.0}$ & $\mathbf{67.9}$ & $72.7$ & $\mathbf{57.8}$\\ 
    &+MEMO  & $69.9$ & $17.0$ & $56.4$ & $97.2$ & $38.7$ & $81.5$ & $25.6$ & $\mathbf{1.4}$ & $74.5$ & $86.8$ & $67.0$ & $72.5$ & $57.4$\\ 
    \hline
    \rowcolor{Gray}
    \multirow{6}{*}{$0.5$} & UNet & $85.5$ & $28.4$ & $72.7$ & $96.6$ & $34.5$ & $80.3$ & $35.2$ & $11.2$ & $85.1$ & $80.4$ & $76.6$ & $74.8$ & $63.4$\\ 
    &+Tent  & $\mathbf{85.8}$ & $27.5$ & $72.4$ & $96.8$ & $33.8$ & $80.3$ & $\mathbf{35.7}$ & $\mathbf{11.8}$ & $\mathbf{85.6}$ & $82.1$ & $78.2$ & $\mathbf{75.0}$ & $\mathbf{63.8}$ \\ 
    &+SAR   & $85.0$ & $28.8$ & $73.2$ & $96.4$ & $\mathbf{36.5}$ & $80.4$ & $32.6$ & $9.6$  & $82.5$ & $79.0$ & $75.0$ & $74.8$ & $62.8$ \\  
    &+FSeg  & $\mathbf{85.8}$ & $27.5$ & $72.4$ & $96.8$ & $33.9$ & $80.3$ & $\mathbf{35.7}$ & $11.7$ & $85.5$ & $82.1$ & $78.3$ & $\mathbf{75.0}$ & $63.7$ \\ 
    &+MEMO  & $85.7$ & $27.7$ & $71.5$ & $\mathbf{96.9}$ & $33.7$ & $79.5$ & $35.4$ & $11.2$ & $85.3$ & $\mathbf{82.4}$ & $78.1$ & $74.6$ & $63.5$ \\ 
    &+SITA  & $85.5$ & $\mathbf{29.8}$ & $\mathbf{74.8}$ & $96.3$ & $36.0$ & $\mathbf{81.8}$ & $34.2$ & $8.0$  & $82.5$ & $81.7$ & $\mathbf{78.4}$ & $74.3$ & $63.6$ \\ 
    \hline
    \rowcolor{Gray}
    \multirow{6}{*}{$0.0$} & UNet & $\mathbf{66.3}$ & $33.3$ & $68.3$ & $94.4$ & $28.2$ & $74.8$ & $28.1$ & $24.9$ & $89.1$ & $44.8$ & $70.7$ & $74.1$ & $58.1$\\ 
    &+Tent  & $63.1$ & $31.5$ & $65.1$ & $95.2$ & $26.9$ & $75.0$ & $28.3$ & $25.0$ & $89.2$ & $46.3$ & $73.0$ & $\mathbf{74.2}$ & $57.7$ \\ 
    &+SAR   & $63.8$ & $33.0$ & $68.0$ & $92.9$ & $\mathbf{31.5}$ & $74.3$ & $28.0$ & $24.8$ & $88.2$ & $43.2$ & $67.6$ & $74.1$ & $57.5$ \\  
    &+FSeg  & $63.3$ & $31.4$ & $65.1$ & $95.2$ & $27.0$ & $75.0$ & $28.2$ & $25.1$ & $89.3$ & $46.3$ & $73.0$ & $\mathbf{74.2}$ & $57.8$ \\ 
    &+MEMO  & $63.9$ & $31.5$ & $66.0$ & $\mathbf{95.5}$ & $27.4$ & $73.4$ & $\mathbf{28.7}$ & $\mathbf{25.4}$ & $\mathbf{89.4}$ & $46.8$ & $72.9$ & $73.9$ & $57.9$ \\ 
    &+SITA  & $59.1$ & $\mathbf{36.8}$ & $\mathbf{72.2}$ & $91.3$ & $\mathbf{31.5}$ & $\mathbf{78.3}$ & $26.1$ & $23.2$ & $85.0$ & $\mathbf{53.1}$ & $\mathbf{76.9}$ & $72.8$ & $\mathbf{58.9}$ \\ 
    \hline
    \end{tabular}
\label{tab:diff1}
\end{sc}
\end{small}
\end{center}
\end{table*}

\begin{table*}[h]
\fontsize{9pt}{9pt}\selectfont
\caption{
\textbf{Same as Table \ref{tab:diff1} but for the Chest X-ray dataset.}}
\vskip 0.15in
\begin{center}
\begin{small}
\begin{sc}
\begin{tabular}{ll|cccccc|c}
    \hline
    $\lambda$ & Method & CHN$\to$MCU & CHN$\to$JSRT & MCU$\to$CHN & MCU$\to$JSRT & JSRT$\to$CHN & JSRT$\to$MCU & Avg.$\uparrow$\\
    \hline
    \rowcolor{Gray}
    \multirow{5}{*}{$1.0$} & Baseline & $\mathbf{86.2}$ & $\mathbf{95.2}$ & $88.2$ & $72.6$ & $92.1$ & $62.8$ & $82.9$ \\ 
    & +Tent & $\mathbf{86.2}$ & $\mathbf{95.2}$ & $\mathbf{88.5}$ & $71.7$ & $\mathbf{92.5}$ & $61.2$ & $82.6$ \\
    & +SAR  & $85.5$ & $95.0$ & $87.5$ & $\mathbf{73.1}$ & $91.0$ & $\mathbf{67.6}$ & $\mathbf{83.3}$ \\
    & +FSeg & $\mathbf{86.2}$ & $\mathbf{95.2}$ & $\mathbf{88.5}$ & $71.7$ & $\mathbf{92.5}$ & $61.3$ & $82.6$ \\
    & +MEMO & $85.0$ & $95.1$ & $88.1$ & $72.6$ & $91.7$ & $60.0$ & $82.1$ \\
    \hline
    \rowcolor{Gray}
    \multirow{6}{*}{$0.5$} & Baseline & $95.2$ & $\mathbf{96.2}$ & $92.5$ & $90.4$ & $93.8$ & $92.6$ & $93.4$ \\ 
    & +Tent & $95.3$ & $\mathbf{96.2}$ & $92.6$ & $\mathbf{90.7}$ & $94.0$ & $\mathbf{92.8}$ & $\mathbf{93.6}$ \\ 
    & +SAR  & $94.8$ & $96.1$ & $92.2$ & $89.6$ & $93.7$ & $92.1$ & $93.1$ \\ 
    & +FSeg & $95.3$ & $\mathbf{96.2}$ & $\mathbf{92.7}$ & $\mathbf{90.7}$ & $\mathbf{94.1}$ & $\mathbf{92.8}$ & $\mathbf{93.6}$ \\ 
    & +MEMO & $95.0$ & $96.1$ & $92.5$ & $90.1$ & $94.0$ & $92.7$ & $93.4$ \\ 
    & +SITA & $\mathbf{95.6}$ & $96.1$ & $92.1$ & $90.5$ & $93.7$ & $92.7$ & $93.5$ \\ 
    \hline 
    \rowcolor{Gray}
    \multirow{6}{*}{$0.0$} & Baseline & $\mathbf{95.6}$ & $95.9$ & $93.0$ & $93.7$ & $90.7$ & $88.9$ & $93.0$ \\ 
    & +Tent & $94.6$ & $\mathbf{96.0}$ & $\mathbf{93.2}$ & $\mathbf{93.8}$ & $\mathbf{91.2}$ & $89.5$ & $93.0$ \\ 
    & +SAR  & $94.3$ & $95.8$ & $92.8$ & $93.5$ & $90.4$ & $88.3$ & $92.5$ \\ 
    & +FSeg & $94.6$ & $\mathbf{96.0}$ & $\mathbf{93.2}$ & $\mathbf{93.8}$ & $\mathbf{91.2}$ & $89.5$ & $\mathbf{93.1}$ \\ 
    & +MEMO & $94.6$ & $95.8$ & $93.1$ & $93.5$ & $\mathbf{91.2}$ & $89.6$ & $93.0$ \\ 
    & +SITA & $94.7$ & $\mathbf{96.0}$ & $92.6$ & $93.7$ & $90.9$ & $\mathbf{90.0}$ & $93.0$ \\ 
    \hline 
\end{tabular}
\end{sc}
\end{small}
\end{center}
\label{tab:diff2}
\end{table*}

\begin{table*}[h]
\fontsize{9pt}{9pt}\selectfont
\centering
\caption{
\textbf{Same as Table \ref{tab:diff1} but for the Retinal Fundus dataset.}}
\vskip 0.15in
\begin{center}
\begin{small}
\begin{sc}
\begin{tabular}{ll|cccccc|c}
    \hline
    $\lambda$ & Method & CHASE$\to$HRF & CHASE$\to$RITE & HRF$\to$CHASE & HRF$\to$RITE & RITE$\to$CHASE & RITE$\to$HRF & Avg.$\uparrow$\\
    \hline
    \rowcolor{Gray}
    \multirow{5}{*}{$1.0$} & Baseline & $52.3$ & $40.5$ & $61.9$ & $53.6$ & $55.5$ & $\mathbf{56.3}$ & $53.3$ \\ 
    & +Tent & $52.1$ & $39.6$ & $61.2$ & $53.1$ & $54.4$ & $\mathbf{56.3}$ & $52.8$ \\
    & +SAR  & $\mathbf{52.5}$ & $\mathbf{41.2}$ & $\mathbf{62.4}$ & $\mathbf{53.9}$ & $\mathbf{56.3}$ & $\mathbf{56.3}$ & $\mathbf{53.8}$ \\
    & +FSeg & $52.1$ & $39.5$ & $61.1$ & $53.1$ & $54.3$ & $\mathbf{56.3}$ & $52.8$ \\
    & +MEMO & $52.1$ & $39.5$ & $61.1$ & $53.0$ & $54.3$ & $\mathbf{56.3}$ & $52.7$ \\
    \hline
    \rowcolor{Gray}
    \multirow{6}{*}{$0.5$} & Baseline & $54.6$ & $55.1$ & $64.0$ & $55.0$ & $67.4$ & $55.8$ & $58.6$ \\ 
    & +Tent & $54.6$ & $54.6$ & $63.6$ & $54.6$ & $67.2$ & $\mathbf{55.9}$ & $58.4$ \\ 
    & +SAR  & $54.6$ & $55.2$ & $64.0$ & $55.1$ & $67.4$ & $55.6$ & $58.7$ \\ 
    & +FSeg & $54.5$ & $54.5$ & $63.6$ & $54.6$ & $67.2$ & $\mathbf{55.9}$ & $58.4$ \\ 
    & +MEMO & $54.5$ & $54.4$ & $63.6$ & $54.5$ & $67.2$ & $\mathbf{55.9}$ & $58.3$ \\ 
    & +SITA & $\mathbf{54.7}$ & $\mathbf{56.0}$ & $\mathbf{64.4}$ & $\mathbf{55.9}$ & $\mathbf{67.5}$ & $55.8$ & $\mathbf{59.0}$ \\ 
    \hline 
    \rowcolor{Gray}
    \multirow{6}{*}{$0.0$} & Baseline & $54.2$ & $60.7$ & $\mathbf{61.4}$ & $55.2$ & $67.6$ & $54.4$ & $58.9$ \\ 
    & +Tent & $\mathbf{54.3}$ & $60.4$ & $\mathbf{61.4}$ & $54.9$ & $\mathbf{68.0}$ & $\mathbf{54.7}$ & $58.9$ \\ 
    & +SAR  & $54.2$ & $60.7$ & $61.2$ & $55.3$ & $67.3$ & $54.2$ & $58.8$ \\ 
    & +FSeg & $\mathbf{54.3}$ & $60.4$ & $\mathbf{61.4}$ & $54.9$ & $\mathbf{68.0}$ & $\mathbf{54.7}$ & $\mathbf{59.0}$ \\ 
    & +MEMO & $\mathbf{54.3}$ & $60.4$ & $\mathbf{61.4}$ & $54.9$ & $\mathbf{68.0}$ & $\mathbf{54.7}$ & $58.9$ \\ 
    & +SITA & $54.2$ & $\mathbf{61.3}$ & $60.3$ & $\mathbf{56.0}$ & $67.2$ & $54.5$ & $58.9$ \\ 
    \hline 
\end{tabular}
\end{sc}
\end{small}
\end{center}

\label{tab:diff3}
\end{table*}

\begin{table*}[t]
\fontsize{8pt}{8pt}\selectfont
\centering
\caption{
{\bf Top:} Baseline UNet performance with different batch norm. statistics $S_{mix}^{\lambda}$, averaged over all domain shifts, for the {\bf Spinal Cord (SC) Dataset}. {\bf Bottom:} Integrated performance of the top block using different integration strategies.}
\vskip 0.15in
\begin{center}
\begin{small}
\begin{sc}
\begin{tabular}{ll|ccccccccccccc}
    \hline
    Method & BN stat. & $1\!\to\!2$ & $1\!\to\!3$ & $1\!\to\!4$ & $2\!\to\!1$ & $2\!\to\!3$ & $2\!\to\!4$ & $3\!\to\!1$ & $3\!\to\!2$ & $3\!\to\!4$ & $4\!\to\!1$ & $4\!\to\!2$ & $4\!\to\!3$ & Avg.$\uparrow$\\
    \hline
    \multirow{5}{*}{UNet} & $\lambda\!=\!0.0$ & $71.2$ & $17.4$ & $58.5$ & $97.2$ & $39.4$ & $82.2$ & $26.2$ & $1.3$ & $74.5$ & $86.0$ & $67.2$ & $72.7$ & $57.8$  \\ 
    & $\lambda\!=\!0.2$ & $84.1$ & $21.6$ & $67.6$ & $97.3$ & $38.1$ & $81.7$ & $30.9$ & $2.9$ & $79.2$ & $84.8$ & $73.2$ & $74.0$ & $61.3$  \\ 
    & $\lambda\!=\!0.4$ & $86.6$ & $26.5$ & $71.8$ & $96.9$ & $35.7$ & $80.9$ & $34.1$ & $7.4$ & $83.3$ & $82.6$ & $76.0$ & $74.7$ & $63.0$  \\ 
    & $\lambda\!=\!0.6$ & $83.2$ & $29.8$ & $72.8$ & $96.3$ & $33.2$ & $79.7$ & $35.1$ & $15.7$ & $86.7$ & $77.4$ & $76.8$ & $74.9$ & $63.5$ \\
    & $\lambda\!=\!0.8$ & $76.4$ & $31.3$ & $71.4$ & $95.6$ & $30.6$ & $77.8$ & $34.0$ & $25.9$ & $88.8$ & $66.8$ & $75.3$ & $74.8$ & $62.4$ \\
    & $\lambda\!=\!1.0$ & $66.3$ & $33.3$ & $68.3$ & $94.4$ & $28.2$ & $74.8$ & $28.1$ & $24.9$ & $89.1$ & $44.8$ & $70.7$ & $74.1$ & $58.1$ \\ 
    \hline
    & Integr. strat. \\
    \hline
    \multirow{6}{*}{InTEnt}& Average & $86.9$ & $26.6$ & $70.9$ & $97.2$ & $34.3$ & $80.1$ & $39.6$ & $13.3$ & $85.5$ & $83.1$ & $78.1$ & $74.8$ & $64.2$ \\
    & Entropy    & $86.9$ & $26.5$ & $70.9$ & $97.2$ & $34.3$ & $80.1$ & $39.7$ & $14.3$ & $85.8$ & $83.5$ & $78.2$ & $74.8$ & $64.3$ \\
    & Ent.-Min   & $64.6$ & $27.7$ & $64.3$ & $97.5$ & $31.5$ & $79.2$ & $29.4$ & $24.4$ & $89.1$ & $83.0$ & $72.1$ & $74.2$ & $61.4$  \\
    & Ent.-Topk  & $84.7$ & $26.1$ & $68.1$ & $97.5$ & $32.9$ & $79.8$ & $37.1$ & $22.8$ & $88.5$ & $84.3$ & $76.5$ & $74.9$ & $64.4$ \\
    & Ent.-Norm  & $85.8$ & $26.4$ & $70.0$ & $97.4$ & $33.2$ & $80.0$ & $38.8$ & $21.1$ & $87.7$ & $84.4$ & $78.4$ & $74.9$ & $64.8$  \\ 
    & Ent.-Baln  & $86.6$ & $28.7$ & $71.4$ & $97.6$ & $35.8$ & $80.7$ & $36.7$ & $12.4$ & $86.5$ & $83.3$ & $79.2$ & $75.0$ & $64.5$  \\
    & Sharpness  & $85.1$ & $27.4$ & $70.8$ & $97.2$ & $35.0$ & $79.5$ & $38.5$ & $15.8$ & $86.1$ & $80.0$ & $78.0$ & $74.9$ & $64.0$  \\ 
    \hline 
\end{tabular}
\end{sc}
\end{small}
\end{center}

\label{tab:internal1}
\end{table*}

\begin{table*}[t]
\fontsize{9pt}{9pt}\selectfont
\centering
\caption{
\textbf{Same as Table \ref{tab:internal1} but for Chest X-Ray Dataset.}
}
\vskip 0.15in
\begin{center}
\begin{small}
\begin{sc}
\begin{tabular}{ll|cccccccccccc}
    \hline
    Method & BN stat. & CHN$\to$MCU & CHN$\to$JSRT & MCU$\to$CHN & MCU$\to$JSRT & JSRT$\to$CHN & JSRT$\to$MCU & Avg.$\uparrow$\\
    \hline
    \multirow{5}{*}{UNet} & $\lambda\!=\!0.0$ & $86.2$ & $95.2$ & $88.2$ & $72.6$ & $92.1$ & $62.8$ & $82.9$ \\ 
    & $\lambda\!=\!0.2$ & $92.4$ & $95.7$ & $90.7$ & $81.4$ & $93.8$ & $88.5$ & $90.4$ \\ 
    & $\lambda\!=\!0.4$ & $94.7$ & $96.1$ & $92.0$ & $88.1$ & $94.0$ & $92.7$ & $92.9$ \\ 
    & $\lambda\!=\!0.6$ & $95.3$ & $96.2$ & $92.8$ & $92.1$ & $93.5$ & $92.0$ & $93.7$ \\ 
    & $\lambda\!=\!0.8$ & $95.1$ & $96.2$ & $93.2$ & $93.9$ & $92.5$ & $90.6$ & $93.6$ \\ 
    & $\lambda\!=\!1.0$ & $95.6$ & $95.9$ & $93.0$ & $93.7$ & $90.7$ & $88.9$ & $93.0$ \\ 
    \hline
    & Integr. strat. \\
    \hline
    \multirow{6}{*}{\textbf{InTEnt}} 
    & Average      & $95.1$ & $96.2$ & $92.7$ & $90.8$ & $93.8$ & $92.8$ & $93.6$ \\ 
    & Entropy      & $95.1$ & $96.2$ & $92.7$ & $91.1$ & $93.9$ & $93.0$ & $93.7$ \\   
    & Ent.-Min          & $93.9$ & $96.1$ & $93.4$ & $93.7$ & $94.1$ & $90.3$ & $93.6$ \\ 
    & Ent.-TopK      & $95.1$ & $96.2$ & $93.3$ & $93.6$ & $94.1$ & $92.9$ & $94.2$ \\
    & Ent.-Norm         & $95.3$ & $96.2$ & $93.1$ & $92.7$ & $94.0$ & $93.4$ & $94.1$ \\
    & Ent.-Baln         & $95.5$ & $96.3$ & $93.1$ & $92.7$ & $94.0$ & $94.3$ & $94.3$ \\
    & Sharpness        & $95.1$ & $96.2$ & $92.8$ & $91.8$ & $93.8$ & $92.8$ & $93.7$ \\ 
    \hline 
\end{tabular}
\end{sc}
\end{small}
\end{center}

\label{tab:internal2}
\end{table*}

\begin{table*}[t]
\fontsize{8pt}{8pt}\selectfont
\centering
\caption{
\textbf{Same as Table \ref{tab:internal1} but for Retinal Fundus Dataset.}
}
\vskip 0.15in
\begin{center}
\begin{small}
\begin{sc}
\begin{tabular}{ll|cccccccccccc}
    \hline
    Method & BN stat. & CHS$\to$HRF & CHS$\to$RITE & HRF$\to$CHS & HRF$\to$RITE & RITE$\to$CHS & RITE$\to$HRF & Avg.$\uparrow$\\
    \hline
    \multirow{5}{*}{UNet} & $\lambda\!=\!0.0$ & $52.3$ & $40.5$ & $61.9$ & $53.6$ & $55.5$ & $56.3$ & $53.3$ \\ 
    & $\lambda\!=\!0.2$ & $53.7$ & $48.1$ & $63.6$ & $54.3$ & $62.7$ & $56.2$ & $56.5$ \\ 
    & $\lambda\!=\!0.4$ & $54.4$ & $53.1$ & $64.0$ & $54.8$ & $66.3$ & $56.1$ & $58.1$ \\ 
    & $\lambda\!=\!0.6$ & $54.7$ & $56.7$ & $63.8$ & $55.1$ & $68.0$ & $55.5$ & $59.0$ \\ 
    & $\lambda\!=\!0.8$ & $54.6$ & $59.2$ & $63.0$ & $55.3$ & $68.2$ & $54.9$ & $59.2$ \\ 
    & $\lambda\!=\!1.0$ & $54.2$ & $60.7$ & $61.4$ & $55.2$ & $67.6$ & $54.4$ & $58.9$ \\ 
    \hline
    & Integr. strat.  \\
    \hline
    \multirow{6}{*}{\textbf{InTEnt}} 
    & Average & $54.6$ & $54.8$ & $64.6$ & $55.1$ & $67.5$ & $56.0$ & $58.8$ \\ 
    & Entropy & $54.6$ & $54.8$ & $64.6$ & $55.1$ & $67.5$ & $56.0$ & $58.8$ \\
    
    & Ent.-Min     & $52.5$ & $53.9$ & $62.7$ & $53.5$ & $56.3$ & $56.3$ & $55.9$ \\ 
    & Ent.-TopK & $53.8$ & $55.1$ & $63.7$ & $54.3$ & $62.7$ & $56.2$ & $57.6$ \\
    & Ent.-Norm    & $54.4$ & $55.1$ & $64.5$ & $54.8$ & $65.9$ & $56.1$ & $58.5$ \\ 
    & Ent.-Baln    & $54.5$ & $53.1$ & $64.5$ & $55.2$ & $68.3$ & $56.1$ & $58.6$ \\ 
    & Sharpness   & $54.7$ & $56.7$ & $64.4$ & $55.0$ & $67.3$ & $56.1$ & $59.0$ \\ 
    \hline 
\end{tabular}
\end{sc}
\end{small}
\end{center}

\label{tab:internal3}
\end{table*}

\end{document}